\documentclass[11pt,a4paper]{article}
\usepackage[hyperref]{acl2018}
\usepackage{times}
\usepackage{latexsym}

\usepackage{amsmath}
\usepackage{amssymb}
\usepackage{color,colortbl}
\usepackage{booktabs}
\usepackage{subcaption}
\usepackage{multirow}
\usepackage{ifthen}
\usepackage{xspace}
\usepackage{graphicx}
\usepackage{framed}
\usepackage{textcomp}
\usepackage{float}
\usepackage{makecell}
\usepackage{booktabs}

\usepackage{url}

\DeclareCaptionFont{10pt}{\fontsize{10pt}{12pt}\selectfont}
\aclfinalcopy

\newcommand\refsec[1]{Section~\ref{sec:#1}}

\newcommand\reffig[1]{Figure~\ref{fig:#1}}

\newcommand\reftab[1]{Table~\ref{tab:#1}}
\newcommand\refapp[1]{Appendix~\ref{sec:#1}}

\ifthenelse{\isundefined{\definition}}{\newtheorem{definition}{Definition}}{}
\ifthenelse{\isundefined{\assumption}}{\newtheorem{assumption}{Assumption}}{}
\ifthenelse{\isundefined{\hypothesis}}{\newtheorem{hypothesis}{Hypothesis}}{}
\ifthenelse{\isundefined{\proposition}}{\newtheorem{proposition}{Proposition}}{}
\ifthenelse{\isundefined{\theorem}}{\newtheorem{theorem}{Theorem}}{}
\ifthenelse{\isundefined{\lemma}}{\newtheorem{lemma}{Lemma}}{}
\ifthenelse{\isundefined{\corollary}}{\newtheorem{corollary}{Corollary}}{}
\ifthenelse{\isundefined{\alg}}{\newtheorem{alg}{Algorithm}}{}
\ifthenelse{\isundefined{\example}}{\newtheorem{example}{Example}}{}
\newcommand{\sneg}{SQuAD~2.0\xspace}
\newcommand{\tfidf}{\textsc{TfIdf}\xspace}
\newcommand{\rulebased}{\textsc{RuleBased}\xspace}
\newcommand\nl[1]{``\textit{#1}''}

\title{Know What You Don't Know: Unanswerable Questions for SQuAD}
\author{
  Pranav Rajpurkar\Thanks{\enskip The first two authors contributed equally to this paper.}
  \qquad Robin Jia$^*$ \qquad Percy Liang \\
  Computer Science Department, Stanford University \\
  {\tt \{pranavsr,robinjia,pliang\}@cs.stanford.edu}
}

\date{}

\begin{document}
\maketitle
\begin{abstract}
  Extractive reading comprehension systems can often
locate the correct answer to a question in a context document,
but they also tend to make unreliable guesses on questions
for which the correct answer is not stated in the context.
Existing datasets either focus exclusively on answerable questions,
or use automatically generated unanswerable questions that are easy to identify.
To address these weaknesses, we present \sneg,
the latest version of the
Stanford Question Answering Dataset (SQuAD).
\sneg combines existing SQuAD data with 
over 50,000 unanswerable questions written adversarially
by crowdworkers to look similar to answerable ones.
To do well on \sneg, systems must not only answer questions when possible,
but also determine when no answer is supported by the paragraph
and abstain from answering.
\sneg is a challenging natural language understanding task for existing models:
a strong neural system that gets
$86\%$ F1 on SQuAD~1.1 achieves only $66\%$ F1 on \sneg.

\end{abstract}

\section{Introduction}
\begin{figure}[t]
  \begin{framed}
  \footnotesize
    \textbf{Article:} Endangered Species Act

    \textbf{Paragraph:}
    \nl{
      \dots Other legislation followed, including the Migratory Bird Conservation Act of 1929, a \textcolor{blue}{1937 treaty} prohibiting the hunting of right and gray whales, and the \textcolor{red}{Bald Eagle Protection Act of 1940}. These \textcolor{red}{later laws} had a low cost to society---the species were relatively rare---and little \textcolor{blue}{opposition} was raised.}

    \vspace{0.08in}

    \textbf{Question 1:} \nl{Which laws
    faced significant \textcolor{blue}{opposition}?}

    \textbf{Plausible Answer:} \textit{\textcolor{red}{later laws}}
    
    \vspace{0.08in}

    \textbf{Question 2:} \nl{What was the name of the \textcolor{blue}{1937 treaty}?}

    \textbf{Plausible Answer:} \textit{\textcolor{red}{Bald Eagle Protection Act}}

  \end{framed}
  \caption{Two unanswerable questions written by crowdworkers,
  along with plausible (but incorrect) answers.
  Relevant keywords are shown in \textcolor{blue}{blue}.
  }
  \label{fig:intro}
\end{figure}

Machine reading comprehension has become a central task in
natural language understanding,
fueled by the creation of many large-scale datasets 
\citep{hermann2015read,hewlett2016wikireading,rajpurkar2016squad,nguyen2016ms,trischler2017newsqa,joshi2017triviaqa}.
In turn, these datasets have spurred
a diverse array of model architecture improvements
\citep{seo2016bidaf,hu2017mnemonic,wang2017gated,clark2017simple,huang2018fusion}.
Recent work has even produced
systems that surpass human-level exact match accuracy
on the Stanford Question Answering Dataset (SQuAD),
one of the most widely-used reading comprehension benchmarks
\citep{rajpurkar2016squad}.

Nonetheless, these systems
are still far from true language understanding.
Recent analysis shows that models can do well at SQuAD by
learning context and type-matching heuristics \citep{weissenborn2017fastqa},
and that success on SQuAD does not ensure robustness
to distracting sentences \citep{jia2017adversarial}.
One root cause of these problems is SQuAD's
focus on questions for which a correct answer is guaranteed to exist
in the context document.
Therefore, models only need to select the span
that seems most related to the question,
instead of checking that the answer is actually entailed by the text.

In this work, we construct \sneg,\footnote{
In the ACL version of this paper, we called our new dataset \textsc{SQuADRUn};
here we use the name \sneg, to emphasize that it is in fact the new version of SQuAD.}
a new dataset that combines answerable questions from the previous version of SQuAD (SQuAD~1.1) with
53,775 new, unanswerable questions about the same paragraphs.
Crowdworkers crafted these questions so that (1) they
are \emph{relevant} to the paragraph,
and (2) the paragraph contains a \emph{plausible answer}---something
of the same type as what the question asks for.
Two such examples are shown in \reffig{intro}.

We confirm that \sneg is both challenging and high-quality.
A state-of-the-art model achieves only $66.3\%$ F1 score when 
trained and tested on \sneg,
whereas human accuracy is $89.5\%$ F1, a full $23.2$ points higher.
The same model architecture trained on SQuAD 1.1 gets $85.8\%$ F1, 
only $5.4$ points worse than humans.
We also show that our unanswerable questions
are more challenging than ones created automatically, 
either via distant supervision \citep{clark2017simple}
or a rule-based method \citep{jia2017adversarial}.
We release \sneg to the public as new version of SQuAD,
and make it the primary benchmark on the official SQuAD leaderboard.\footnote{
  As with previous versions of SQuAD, we release \sneg under 
  the CC BY-SA 4.0 license.}
We are optimistic that this new dataset will encourage the development of reading 
comprehension systems that know what they don't know.

\section{Desiderata}
\label{sec:desiderata}
We first outline our goals for \sneg.
Besides the generic goals of
large size, diversity, and low noise,
we posit two desiderata specific to unanswerable questions:

\paragraph{Relevance.} The unanswerable questions
should appear relevant to the topic of the
context paragraph.
Otherwise, simple heuristics (e.g., based on word overlap)
could distinguish answerable and unanswerable questions
\citep{yih2013enhanced}.

\paragraph{Existence of plausible answers.}
There should be some span in the context whose type
matches the type of answer the question asks for.
For example, if the question asks,
\nl{What company was founded in 1992?},
then some company should appear in the context.
Otherwise, type-matching heuristics
could distinguish answerable and unanswerable questions
\citep{weissenborn2017fastqa}.

\section{Existing datasets}
Next, we survey existing reading comprehension datasets
with these criteria in mind.
We use the term ``negative example''
to refer to a context passage paired with an unanswerable question.

\subsection{Extractive datasets}
In extractive reading comprehension datasets,
a system must extract the correct answer to a question
from a context document or paragraph.
The Zero-shot Relation Extraction dataset \citep{levy2017zero} 
contains negative examples generated with distant supervision.
\citet{levy2017zero} found that $65\%$ 
of these negative examples do not have a plausible answer, making them 
easy to identify.

Other distant supervision strategies can also create negative examples.
TriviaQA \citep{joshi2017triviaqa} 
retrieves context documents from the web or Wikipedia for each question.
Some documents do not contain the correct answer,
yielding negative examples;
however, these are excluded from the final dataset.
\citet{clark2017simple} generate negative examples
for SQuAD by pairing existing questions with other paragraphs
from the same article based on TF-IDF overlap;
we refer to these as \tfidf examples.
In general, distant supervision does not ensure the existence of a
plausible answer in the retrieved context,
and might also add noise, as the context might
contain a paraphrase of the correct answer.
Moreover, when retrieving from a small set of possible contexts,
as in \citet{clark2017simple},
we find that the retrieved paragraphs are often not very
relevant to the question, making these negative examples
easy to identify.

The NewsQA data collection process also yields unanswerable questions,
because crowdworkers write questions given only a summary of an article,
not the full text \citep{trischler2017newsqa}.
Only $9.5\%$ of their questions are unanswerable,
making this strategy hard to scale.
Of this fraction, we found that some are misannotated as unanswerable,
and others are out-of-scope (e.g., summarization questions).
\citet{trischler2017newsqa} also exclude negative examples
from their final dataset.

\citet{jia2017adversarial} propose a rule-based procedure
for editing SQuAD questions to make them unanswerable.
Their questions are not very diverse:
they only replace entities and numbers with similar words, 
and replace nouns and adjectives with WordNet antonyms.
We refer to these unanswerable questions as \rulebased questions.

\begin{table*}[th]
  \footnotesize
  \centering
  \begin{tabular}{|lllr|}
    \hline
    Reasoning & Description & Example & Percentage \\
    \hline
    Negation  
    & \makecell[l]{Negation word inserted \\ or removed.}
    & \makecell[l]{
      Sentence: 
      ``\textit{Several hospital pharmacies have decided to}\\
        \textit{outsource high risk preparations \dots}'' \\
      Question: 
      ``\textit{What types of pharmacy functions have \textbf{never}}\\
        \textit{been outsourced?}''}
    & $9\%$ \\
    \hline
    Antonym  & \makecell[l]{Antonym used.}
    & \makecell[l]{
      S: 
      ``\textit{the extinction of the dinosaurs\dots allowed the} \\
        \textit{tropical rainforest to spread out across the continent.}'' \\
      Q: 
      ``\textit{The extinction of what led to the \textbf{decline} of rainforests?}''}
    & $20\%$ \\
    \hline
    \makecell[l]{Entity Swap} & \makecell[l]{
      Entity, number, or date
      \\ replaced with other 
      \\ entity, number, or date.}
    & \makecell[l]{
      S: 
      ``\textit{These values are much greater than the 9--88 cm}\\
        \textit{as projected \dots in its Third Assessment Report.}''\\
      Q: 
      ``\textit{What was the projection of sea level increases in the }\\
        \textit{\textbf{fourth assessment report}?}''}
    & $21\%$ \\
    \hline
    \makecell[l]{Mutual \\ Exclusion} & \makecell[l]{
      Word or phrase is \\
      mutually exclusive \\
      with something for which \\
      an answer is present.}
    & \makecell[l]{
      S: 
      ``\textit{BSkyB\dots waiv[ed] the charge for subscribers whose} \\
        \textit{package included two or more premium channels.}'' \\
      Q: 
      ``\textit{What service did BSkyB \textbf{give away for free}} \\
        \textit{\textbf{unconditionally}?}''}
    & $15\%$ \\
    \hline
    \makecell[l]{Impossible \\ Condition} & \makecell[l]{
      Asks for condition that \\
      is not satisfied by \\
      anything in the paragraph.}
    & \makecell[l]{
      S: 
      ``\textit{Union forces left Jacksonville and confronted}\\
        \textit{a Confederate Army at the Battle of Olustee\dots}\\
        \textit{Union forces then retreated to Jacksonville}\\
        \textit{and held the city for the remainder of the war.}''\\
      Q: 
      ``\textit{After what battle did Union forces leave}\\
        \textit{Jacksonville \textbf{for good}?}''}
    & $4\%$ \\
    \hline
    \makecell[l]{Other \\ Neutral} & \makecell[l]{
      Other cases where the \\
      paragraph does not imply \\
      any answer.}
    & \makecell[l]{
      S: 
      ``\textit{Schuenemann et al. concluded in 2011 that the}\\
        \textit{Black Death\dots was caused by a variant of Y. pestis\dots}''\\
      Q: 
      ``\textit{Who \textbf{discovered} Y. pestis?}''}
    & $24\%$ \\
    \hline
    Answerable & \makecell[l]{Question is answerable \\ (i.e. dataset noise).}
    & & $7\%$ \\
    \hline
  \end{tabular}
  \caption{Types of negative examples in \sneg exhibiting a wide range of phenomena.}
  \label{tab:phenomena}
\end{table*}

\subsection{Answer sentence selection datasets}
Sentence selection datasets test whether a system can rank
sentences that answer a question higher than sentences that do not.
\citet{wang2007qa} constructed the \textsc{QASent} dataset
from questions in the TREC 8-13 QA tracks.
\citet{yih2013enhanced} showed that
lexical baselines are highly competitive on this dataset.
WikiQA \citep{yang2015wikiqa} pairs questions from Bing query logs
with sentences from Wikipedia.
Like \tfidf examples,
these sentences are not guaranteed to have plausible answers
or high relevance to the question.
The dataset is also limited in scale (3,047 questions, 1,473 answers).

\subsection{Multiple choice datasets}
Finally, some datasets, like
MCTest \citep{richardson2013mctest}
and RACE \citep{lai2017race},
pose multiple choice questions,
which can have a ``none of the above'' option.
In practice, multiple choice options are often unavailable,
making these datasets less suited for training user-facing systems.
Multiple choice questions
also tend to be quite different from extractive ones,
with more emphasis on
fill-in-the-blank, interpretation, and summarization
\citep{lai2017race}.

\section{\sneg}
We now describe our new dataset,
which we constructed to satisfy both the relevance and plausible answer
desiderata from \refsec{desiderata}.

\subsection{Dataset creation}
We employed crowdworkers
on the Daemo crowdsourcing platform \citep{gaikwad2015daemo}
to write unanswerable questions.
Each task consisted of an entire article from SQuAD 1.1.
For each paragraph in the article,
workers were asked to pose up to five questions that were
impossible to answer based on the paragraph alone, 
while referencing entities in the paragraph
and ensuring that a plausible answer is present.
As inspiration,
we also showed questions from SQuAD~1.1 for each paragraph;
this further encouraged unanswerable questions to look
similar to answerable ones.
Workers were asked to spend 7 minutes per paragraph, 
and were paid \$10.50 per hour. 
Screenshots of our interface are shown in \refapp{interface}.

We removed questions from workers
who wrote 25 or fewer questions on that article;
this filter helped remove noise from workers who had trouble
understanding the task, and therefore quit before completing
the whole article.
We applied this filter to both our new data and the
existing answerable questions from SQuAD~1.1.
To generate train, development, and test splits,
we used the same partition of articles as SQuAD 1.1,
and combined the existing data with our new data for each split.
For the \sneg development and test sets, we removed
articles for which we did not collect unanswerable questions. 
This resulted in a roughly one-to-one ratio of answerable to unanswerable
questions in these splits,
whereas the train data has roughly twice as many answerable questions 
as unanswerable ones.
\reftab{stats} summarizes overall statistics of \sneg.

\begin{table}[t]
  \footnotesize
  \centering
  \begin{tabular}{|l|r|r|}
    \hline
    & SQuAD 1.1 & \sneg \\
    \hline
    \textbf{Train}          &        &         \\
    Total examples          & 87,599 & 130,319 \\
    Negative examples       &      0 &  43,498 \\
    Total articles          &    442 &     442 \\
    Articles with negatives &      0 &     285 \\
    \hline
    \textbf{Development}    &        &         \\
    Total examples          & 10,570 &  11,873 \\
    Negative examples       &      0 &   5,945 \\
    Total articles          &     48 &      35 \\
    Articles with negatives &      0 &      35 \\
    \hline
    \textbf{Test}           &        &         \\
    Total examples          &  9,533 &   8,862 \\
    Negative examples       &      0 &   4,332 \\
    Total articles          &     46 &      28 \\
    Articles with negatives &      0 &      28 \\
    \hline
  \end{tabular}
  \caption{Dataset statistics of \sneg, compared to the previous SQuAD 1.1.}
  \label{tab:stats}
\end{table}

\begin{table*}[th]
  \footnotesize
  \centering
  \begin{tabular}{|l|cc|cc|cc|}
    \hline
    \multirow{2}{*}{System}
    & \multicolumn{2}{c|}{SQuAD 1.1 test}
    & \multicolumn{2}{c|}{\sneg dev}
    & \multicolumn{2}{c|}{\sneg test} \\
    & EM & F1 & EM & F1 & EM & F1\\
    \hline
    BNA          & $68.0$ & $77.3$ & $59.8$ & $62.6$ & $59.2$ & $62.1$ \\
    DocQA        & $72.1$ & $81.0$ & $61.9$ & $64.8$ & $59.3$ & $62.3$ \\
    DocQA + ELMo & $\bf 78.6$ & $\bf 85.8$ & $\bf 65.1$ & $\bf 67.6$ 
                 & $\bf 63.4$ & $\bf 66.3$ \\
    \hline
    Human        & $82.3$ & $91.2$ & $86.3$ & $89.0$ & $86.9$ & $89.5$ \\
    Human--Machine Gap 
                 & $3.7$ & $5.4$ & $\bf 21.2$ & $\bf 21.4$
                 & $\bf 23.5$ & $\bf 23.2$ \\
    \hline
  \end{tabular}
  \caption{Exact Match (EM) and F1 scores on SQuAD 1.1 and 2.0.
  The gap between humans and the best tested model is much
  larger on \sneg,
  suggesting there is a great deal of room for model improvement.
  }
  \label{tab:results}
\end{table*}
\begin{table*}[th]
  \footnotesize
  \centering
  \begin{tabular}{|l|cc|cc|cc|}
    \hline
    \multirow{2}{*}{System}
    & \multicolumn{2}{c|}{SQuAD 1.1 + \tfidf}
    & \multicolumn{2}{c|}{SQuAD 1.1 + \rulebased}
    & \multicolumn{2}{c|}{\sneg dev} \\
    & EM & F1 & EM & F1 & EM & F1\\
    \hline
    BNA          & $72.7$ & $76.6$ & $80.1$ & $84.8$ & $59.8$ & $62.6$ \\
    DocQA        & $75.6$ & $79.2$ & $80.8$ & $84.8$ & $61.9$ & $64.8$ \\
    DocQA + ELMo & $\bf 79.4$ & $\bf 83.0$ & $\bf 85.7$ & $\bf 89.6$ 
                 & $\bf 65.1$ & $\bf 67.6$ \\
    \hline
  \end{tabular}
  \caption{Exact Match (EM) and F1 scores on the \sneg development set,
  compared with SQuAD 1.1 with two types of automatically generated negative examples.  
  \sneg is more challenging for current models.
  }
  \label{tab:other}
\end{table*}

\subsection{Human accuracy}
To confirm that our dataset is clean,
we hired additional crowdworkers to answer all questions in the 
\sneg development and test sets.
In each task, we showed workers an entire article from the dataset.
For each paragraph, we showed all associated questions;
unanswerable and answerable questions were shuffled together.
For each question, workers were told to either highlight the answer 
in the paragraph, or mark it as unanswerable. 
Workers were told to expect every paragraph to have some 
answerable and some unanswerable questions.
They were asked to spend one minute per question, 
and were paid \$10.50 per hour.

To reduce crowdworker noise,
we collected multiple human answers for each question
and selected the final answer by majority vote,
breaking ties in favor of answering questions
and preferring shorter answers to longer ones.
On average, we collected 4.8 answers per question.
We note that for SQuAD 1.1,
\citet{rajpurkar2016squad} evaluated a single human's performance;
therefore, they likely underestimate human accuracy.

\subsection{Analysis}
We manually inspected 100 randomly chosen negative examples
from our development set to understand the challenges
these examples present.
In \reftab{phenomena}, we define different categories of
negative examples, and give examples and their frequency in \sneg.
We observe a wide range of phenomena, extending beyond
expected phenomena like negation, antonymy, and entity changes.
In particular, \sneg is much more diverse than \rulebased,
which creates unanswerable questions by applying entity, number, and antonym
swaps to existing SQuAD 1.1 questions.
We also found that $93\%$ of the sampled negative examples are indeed unanswerable.

\section{Experiments}
\label{sec:experiments}
\subsection{Models}
We evaluated three existing model architectures:
the BiDAF-No-Answer (BNA) model proposed by \citet{levy2017zero},
and two versions of the DocumentQA No-Answer (DocQA) model
from \citet{clark2017simple}, namely versions
with and without ELMo \citep{peters2018elmo}.
These models all learn to predict the probability that a question is unanswerable,
in addition to a distribution over answer choices.
At test time, models abstain whenever their predicted probability that 
a question is unanswerable exceeds some threshold.
We tune this threshold separately for each model 
on the development set.
When evaluating on the test set,
we use the threshold that maximizes F1 score on the development set.
We find this strategy does slightly better than simply taking
the argmax prediction, possibly due to the different
proportions of negative examples at training and test time.

\subsection{Main results}
First, we trained and tested all three models on \sneg,
as shown in \reftab{results}.
Following \citet{rajpurkar2016squad},
we report average exact match and F1 scores.\footnote{
  For negative examples, abstaining receives 
a score of $1$, and any other response gets $0$,
for both exact match and F1.}
The best model, DocQA + ELMo, achieves only $66.3$ F1 on the test set,
$23.2$ points lower than the human accuracy of $89.5$ F1.
Note that a baseline that always abstains gets $48.9$ test F1; 
existing models are closer to this baseline
than they are to human performance.
Therefore, we see significant room for model improvement on this task.
We also compare with reported test numbers
for analogous model architectures on SQuAD 1.1.
There is a much larger gap between humans and machines
on \sneg compared to SQuAD 1.1,
which confirms that \sneg is a much harder dataset
for existing models.

\subsection{Automatically generated negatives}
Next, we investigated whether automatic ways of generating negative examples
can also yield a challenging dataset.
We trained and tested all three model architectures on SQuAD 1.1
augmented with either \tfidf or \rulebased examples.
To ensure a fair comparison with \sneg,
we generated training data by applying \tfidf or \rulebased
only to the 285 articles for which \sneg has unanswerable questions.
We tested on the articles and answerable questions
in the \sneg development set,
adding unanswerable questions in a roughly one-to-one ratio
with answerable ones.
These results are shown in \reftab{other}.
The highest score on \sneg 
is $15.4$ F1 points lower than the highest score on either 
of the other two datasets, suggesting that automatically generated
negative examples are much easier for existing models to detect.

\subsection{Plausible answers as distractors}
\label{sec:plausible}
Finally, we measured how often systems were fooled into answering the plausible
but incorrect answers provided by 
crowdworkers for our unanswerable questions. 
For both computer systems and humans,
roughly half of all wrong answers on unanswerable questions
exactly matched the plausible answers.
This suggests that the plausible answers do indeed serve as
effective distractors.
Full results are shown in \refapp{plausible-supp}.

\section{Discussion}
\sneg forces models to understand whether a paragraph
entails that a certain span is the answer to a question.
Similarly, recognizing textual entailment (RTE)
requires systems to decide whether a hypothesis
is entailed by, contradicted by, or neutral with respect to a 
premise \citep{marelli2014sick,bowman2015large}.
Relation extraction systems must understand
when a possible relationship between two entities
is not entailed by the text \citep{zhang2017tacred}.

\citet{jia2017adversarial} created adversarial test examples
that fool models trained on SQuAD 1.1.
However, models that are trained on similar examples
are not easily fooled by their method.
In contrast, the adversarial examples in \sneg
are difficult even for models trained on examples from the same distribution.

In conclusion, we have presented \sneg, a challenging, diverse, and large-scale 
dataset that forces models to understand when a question
cannot be answered given the context.
We are optimistic that \sneg will encourage the development of new 
reading comprehension models that know what they don't know, 
and therefore understand language at a deeper level.

\paragraph{Reproducibility.} All code, data, experiments are available 
on the CodaLab platform at \url{https://bit.ly/2rDHBgY}.

\paragraph{Acknowledgments.} We would like to thank the anonymous reviewers, 
Arun Chaganty, Peng Qi, and Sharon Zhou for their constructive feedback.
We are grateful to Durim Morina and Michael Bernstein for their 
help with the Daemo platform.
This work was supported by funding from Facebook.
R.J. is supported by an NSF Graduate Research
Fellowship under Grant No. DGE-114747.

\bibliographystyle{acl_natbib}
\bibliography{refdb/all}

\clearpage
\appendix
\section{Supplementary material}
\subsection{Crowdsourcing details}
\label{sec:interface}
\begin{figure*}[t]
  \begin{center}
    \includegraphics[width=\textwidth]{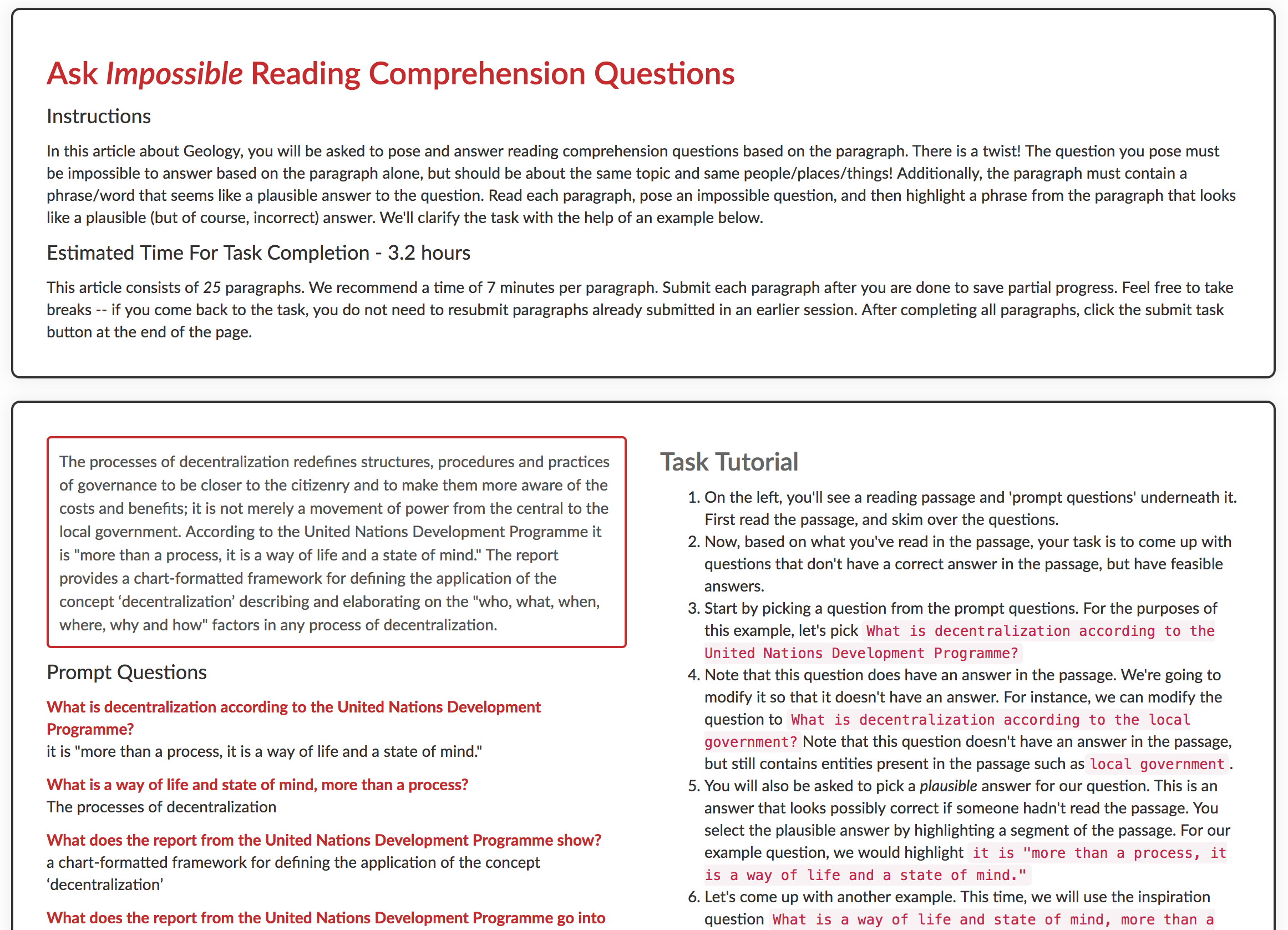}
  \end{center}
  \caption{The instructions shown to crowdworkers
  at the beginning of each question writing task.}
  \label{fig:instructions}
\end{figure*}

\begin{figure*}[t]
  \begin{center}
    \includegraphics[width=\textwidth]{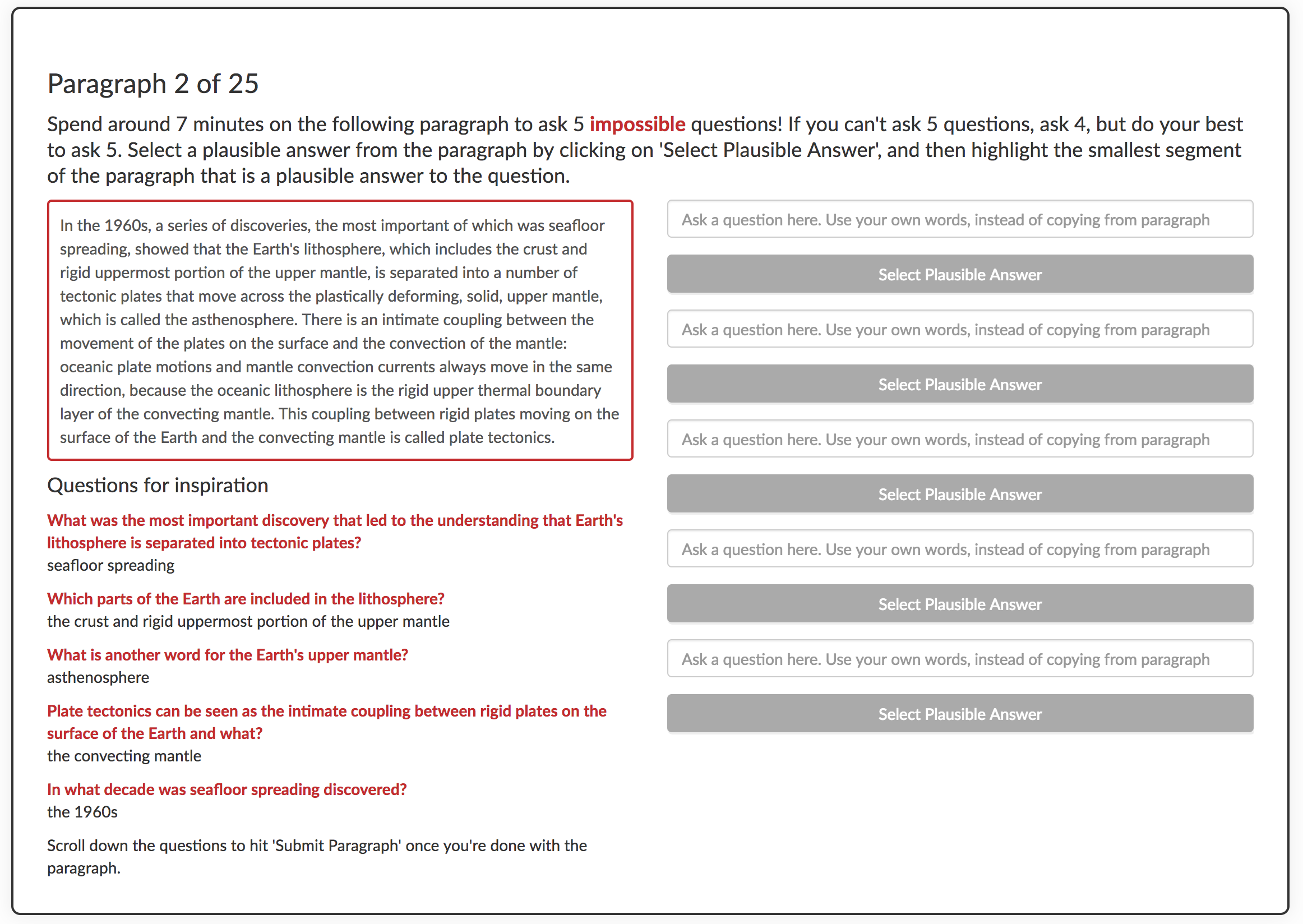}
  \end{center}
  \caption{The interface crowdworkers used to write unanswerable
  questions and annotate plausible answers.}
  \label{fig:interface}
\end{figure*}

\reffig{instructions} shows the instructions that crowdworkers
were given at the beginning of each question writing task.
\reffig{interface} shows the interface they used to write
unanswerable questions for each paragraph.
In the interface, workers first write an unanswerable
question, then highlight a plausible answer in the paragraph.


\subsection{Plausible answers as distractors}
\label{sec:plausible-supp}
As mentioned in \refsec{plausible},
we measured how often systems were fooled into answering
the plausible answers provided by crowdworkers for 
our unanswerable questions.
For each system, we first isolated their 
false positive errors---cases where they predicted an answer
to an unanswerable question---on the development set.
Within this set of examples, we measured exact match and F1 scores
between the system predictions and plausible answers.
These numbers are shown in \reftab{plausible}.
Plausible answers account for roughly half of the false positive errors
made by each of the computer systems, as well as by human answerers.
We conclude that the plausible answers in our dataset
do indeed serve their purpose of being distracting spans
that could be mistaken for the correct answer.

\begin{table}[th]
  \centering
  \begin{tabular}{|l|rr|}
    \hline
    & \multicolumn{1}{c}{EM} & \multicolumn{1}{c|}{F1} \\
    \hline
    BNA         & 48.6 & 63.0 \\
    DocQA       & 55.0 & 69.9 \\
    DocQA +ELMo & 54.9 & 69.2 \\
    \hline
    Human       & 46.4 & 60.6 \\
    \hline
  \end{tabular}
  \caption{Exact match (EM) and F1 scores between system predictions
  and plausible answers, in cases where the system made
  a false positive error.}
  \label{tab:plausible}
\end{table}

\end{document}